%% file: main.tex
\documentclass[a4paper,twoside]{article}

\usepackage{epsfig}
\usepackage{subcaption}
\usepackage{calc}
\usepackage{amssymb}
\usepackage{amstext}
\usepackage{amsmath}
\usepackage{amsthm}
\usepackage{multicol}
\usepackage{pslatex}
\usepackage{apalike}
\usepackage{xcolor}
\usepackage{algorithm2e}
\usepackage{tabularx}
\usepackage{array}
\usepackage[bottom]{footmisc}
\usepackage{adjustbox}
\usepackage{array}
\usepackage{SCITEPRESS}     % Please add other packages that you may need BEFORE the SCITEPRESS.sty package.
\usepackage{multirow}
\usepackage{float}
\usepackage{cuted}
\usepackage{capt-of}
\usepackage{xcolor}

\begin{document}

\title{Exploring Multimodal LMMs for Online \\Episodic Memory Question Answering on the Edge}

\author{\authorname{
Giuseppe Lando\sup{1},
Rosario Forte\sup{1},
Antonino Furnari\sup{1}
}
\affiliation{\sup{1}Department of Mathematics and Computer Science, University of Catania, Italy}
\email{giuseppe.lando@studium.unict.it, rosario.forte@phd.unict.it, antonino.furnari@unict.it}
}

%\title{Exploring Lightweight Multimodal LLMs for Privacy-Preserving, Real-Time Online Episodic Memory Question Answering on the Edge}

%How Far Can MLLMs be pushed on edge devices for streaming episodic memory question answering

% \author{\authorname{First Author Name\sup{1}\orcidAuthor{0000-0000-0000-0000}, Second Author Name\sup{1}\orcidAuthor{0000-0000-0000-0000} and Third Author Name\sup{2}\orcidAuthor{0000-0000-0000-0000}}
% \affiliation{\sup{1}Institute of Problem Solving, XYZ University, My Street, MyTown, MyCountry}
% \affiliation{\sup{2}Department of Computing, Main University, MySecondTown, MyCountry}
% \email{\{first\_author, second\_author\}@ips.xyz.edu, third\_author@dc.mu.edu}
% }


\abstract{We investigate the feasibility of using Multimodal Large Language Models (MLLMs) for real-time online episodic memory question answering. While cloud offloading is common, it raises privacy and latency concerns for wearable assistants, hence we investigate implementation on the edge. We integrated streaming constraints into our question answering pipeline, which is structured into two asynchronous threads: a Descriptor Thread that continuously converts video into a lightweight textual memory, and a Question Answering (QA) Thread that reasons over the textual memory to answer queries. Experiments on the QAEgo4D-Closed benchmark analyze the performance of Multimodal Large Language Models (MLLMs) within strict resource boundaries, showing promising results also when compared to clound-based solutions. Specifically, an end-to-end configuration running on a consumer-grade 8GB GPU achieves 51.76\% accuracy with a Time-To-First-Token (TTFT) of 0.41s. Scaling to a local enterprise-grade server yields 54.40\% accuracy with a TTFT of 0.88s. In comparison, a cloud-based solution obtains an accuracy of 56.00\%. These competitive results highlight the potential of edge-based solutions for privacy-preserving episodic memory retrieval.}

\onecolumn \maketitle \normalsize \setcounter{footnote}{0} \vfill
% \maketitle
% \vspace{-8mm}
% \normalsize
% \setcounter{footnote}{0}

\input{Parts/Introduction}

\input{Parts/Related_Works}

\input{Parts/Method}

\input{Parts/Results}

\input{Parts/Conclusion}

\bibliographystyle{apalike}
{\small
\bibliography{references.bib}}

\end{document}

%% file: Parts/Introduction.tex
\section{\uppercase{Introduction}}
\label{sec:introduction}

The advent of long-form egocentric video datasets such as Ego4D~\cite{grauman2022ego4dworld3000hours} has brought attention to the problem of \emph{episodic memory retrieval}, which is formulated, in one of its variants, as an egocentric Video Question Answering (VideoQA) problem. In particular, the Natural Language Queries (NLQ) task challenges models to retrieve relevant segments from long first-person videos based on natural language questions, requiring both fine-grained temporal localization and multi-modal reasoning over long temporal horizons. While initially defined in an \emph{offline} setting where the entire video is available at query time~\cite{grauman2022ego4dworld3000hours,cvpr24_groundvqa}, such solutions incur storage and computational costs that grow linearly with video length, making them ill-suited for realistic streaming scenarios.

% \begin{figure}[t]
%     \centering
%     \includegraphics[width=\columnwidth]{Images/task.jpeg} % Replace with your file
%     \caption{This four-panel vignette illustrates a common scenario of memory failure caused by distraction or mild cognitive impairment, demonstrating how smart glasses can serve as an episodic memory aid.
%     Panel 1 (Encoding): A man, while distracted, absentmindedly places his car keys inside a drawer.
%     Panel 2 (The Gap): Later, the man wants to come back home, but he forgot the keys to open his car. During this time, his memory of where he placed the keys fades due to the shift in focus.
%     Panel 3 (Retrieval): Realizing the keys are missing, the user consults his smart glasses. The device accesses its episodic memory logs in order to reply to the query.
%     Panel 4 (Resolution): Guided by the technology, the man retrieves the keys from the drawer, resolving the anxiety of the lost item.}
    
%     \label{fig:Task_introduction}
% \end{figure}

\begin{figure*}[t]
    \centering
    \includegraphics[width=\textwidth]{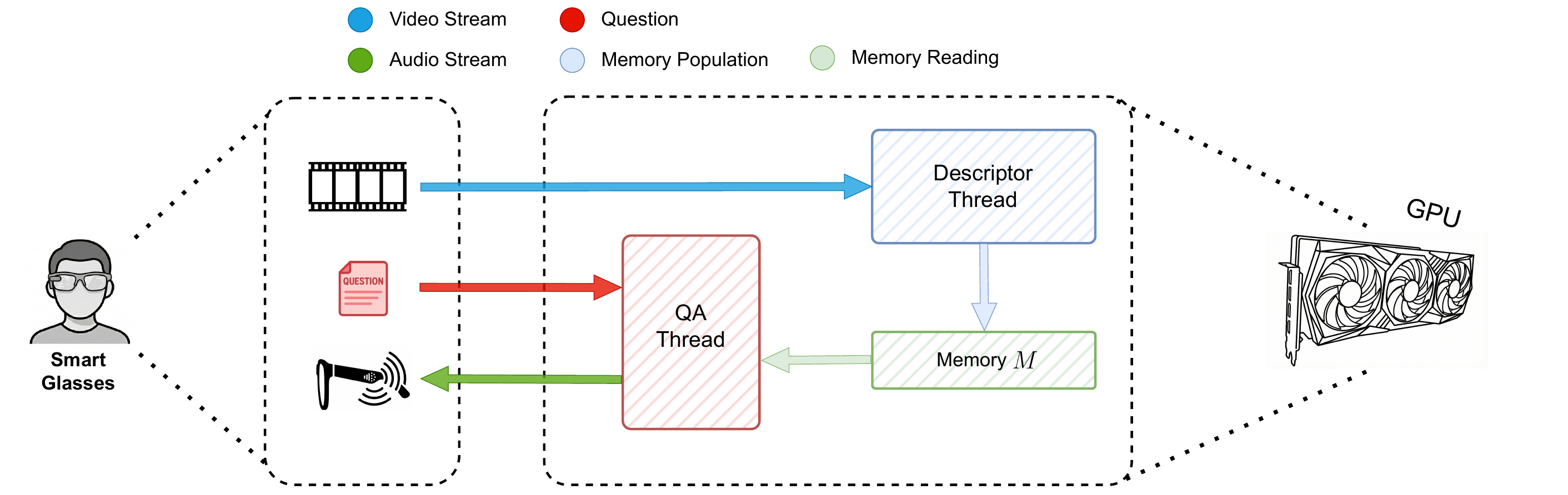} % Replace with your file
    \caption{\textbf{High-level overview of the proposed Edge-based OEM-VQA system.} The user wears smart glasses continuously streaming video to a local unit (GPU). The video is continuously processed into a textual memory M by the Descriptor Thread, allowing the user to ask questions that are ingested into the QA Thread that leverages the textual memory to reply to the user, sending the answer without storing raw video frames.}
    \label{fig:Introductive_architecture}
\end{figure*}
Concurrently, the emergence of Multimodal Large Language Models (MLLMs) has revolutionized video understanding, demonstrating impressive zero-shot capabilities across various tasks~\cite{wang2024internvideo2scalingfoundationmodels,li2024llava,zhang2025videollama3frontiermultimodal,geminiteam2024geminifamilyhighlycapable}. However, these models typically operate in offline settings with high inference latency, further hindering their applicability in real-time streaming scenarios.
To address these limitations, recent research has shifted towards \emph{Online Video Question Answering} (Online VQA), where the system processes video in a streaming fashion without prior knowledge of the question~\cite{di2025streamingvideoquestionansweringincontext,sun2025videosalmonnsstreamingaudiovisual,xu2025streamingvlmrealtimeunderstandinginfinite}. This setting is particularly relevant for wearable assistants and egocentric life-logging systems, where latency and resource constraints play a central role.%: video-language models such as InternVideo2, LLaVA-based systems and Video-LLaMA variants~\cite{wang2024internvideo2scalingfoundationmodels,li2024llava,zhang2025videollama3frontiermultimodal,geminiteam2024geminifamilyhighlycapable} have shown strong zero-shot performance on a variety of video tasks, including VideoQA, by combining visual encoders with powerful language backbones.

Motivated by these advances, recent work has investigated Online Episodic Memory Video Question Answering (OEM-VQA), assessing whether \emph{training-free} approaches can tackle the task without any additional training~\cite{di2025streamingvideoquestionansweringincontext,lando2025oemvqa}.
% In particular~\cite{shen2024encode,wang2024lifelongmemory,yang2025egolifeegocentriclifeassistant,lando2025oemvqa} proposed a training-free pipeline in which an MLLM is used to convert short video clips into a lightweight textual memory, which a language model then reads to answer multiple-choice questions. Despite its simplicity, this paradigm attains competitive accuracy on the QAEgo4D-Closed benchmark~\cite{cvpr24_groundvqa} while requiring only a few kilobytes of storage per minute of video, suggesting that much of the information needed for OEM-VQA can be captured in compact textual form.
%In particular, a line of recent works has explored \emph{training-free} approaches to Online Episodic Memory and egocentric VideoQA based on Multimodal Large Language Models~\cite{shen2024encode,wang2024lifelongmemory,lando2025oemvqa}. 
Notably, recent work~\cite{shen2024encode,wang2024lifelongmemory,lando2025oemvqa} demonstrates that clip-level textual memories are sufficient to achieve competitive accuracy on the QAEgo4D-Closed benchmark~\cite{cvpr24_groundvqa}, while requiring only a few kilobytes of storage per minute of video. 
%This suggests that much of the information required for OEM-VQA can be preserved in highly compact, human-readable memories, without retaining raw visual data.
However, many current deployments of video assistants implicitly assume offloading computation to the cloud, which involves uploading raw frames for storage and inference, trading latency and privacy for convenience. In several assistive scenarios this assumption is not acceptable—e.g., home monitoring or clinical contexts involving cognitive impairment—where regulations, consent, and user trust can prohibit sending first-person video to remote servers. This motivates a strictly \emph{privacy-preserving} regime in which raw video never leaves local infrastructure and only a lightweight textual memory is retained. In this work, we therefore investigate what performance is achievable \emph{without the cloud}, under realistic streaming constraints.
% However, the state-of-the-art configurations identified in~\cite{lando2025oemvqa} depend heavily on large-scale MLLMs or cloud-based, closed-source APIs. Such dependencies render them impractical for real-world deployment in sensitive settings like home assistance, which necessitate strict privacy preservation and operation under limited computational budgets.
%However, the best-performing configurations in~\cite{lando2025oemvqa} rely on \emph{large} multimodal and language models, whose inference latency often exceeds the duration of the processed clips. As a result, they are not suitable for truly real-time, streaming operation. Moreover, most existing OEM-VQA systems implicitly assume access to high-end research hardware and do not explicitly address deployment under tight resource and privacy constraints. In many realistic use cases --- such as home assistants, hospitals or assisted living facilities --- continuous egocentric video is highly sensitive from a privacy perspective, and \emph{cloud offloading is not acceptable}. Computation must therefore be performed \emph{on the edge}, either on a consumer-level device (e.g., a local workstation or embedded GPU) or on an on-premise server within the same institution, without sending raw video to external providers.
These considerations raise a central research question:
\begin{quote}
    \emph{Can multimodal large language models support real-time OEM-VQA on edge hardware, while maintaining competitive accuracy and respecting privacy constraints?}
\end{quote}
In this paper, we investigate this question by exploring the limits of \emph{lightweight} MLLMs in the OEM-VQA setting under strict streaming and deployment constraints. We build upon the architectural paradigm recently investigated in~\cite{shen2024encode,wang2024lifelongmemory,yang2025egolifeegocentriclifeassistant,lando2025oemvqa}, which consists in converting an egocentric video stream into a sequence of textual memory entries—but explicitly focus on \emph{resource-aware} local deployments, spanning a consumer-grade edge device and a more capable on-premise local server. An overview of our proposed framework is illustrated in Figure~\ref{fig:Introductive_architecture}. As shown in the figure, the system processes the video stream entirely on a local unit (e.g., a workstation or a server connected to smart glasses via WiFi), ensuring that raw visual data never leaves the local infrastructure. The video stream is transformed into textual descriptions by a dedicated Descriptor Thread; in parallel, when a query occurs, a QA Thread reads the accumulated memory and uses it as context to formulate and send the answer back to the smart glasses. This decoupled design enables the continuous processing of input clips while keeping the model available to respond to user queries instantly. Concretely, we instantiate this study using different variants of the Qwen3-VL model~\cite{bai2025qwen3vltechnicalreport}, while keeping the overall pipeline training-free.
We analyze performance in two representative deployment regimes: (i) an \emph{edge setting} in which all computations are performed on a single consumer-grade GPU, and (ii) an \emph{enterprise-grade} setting in which inference runs on a more capable local server, strictly without involving cloud resources.
Across both regimes, we explore a range of configurations by varying frame rate, input resolution, batch size, and model size. We enforce a strict \emph{streaming constraint}, requiring that the time needed to generate a textual description of a clip be lower than the clip duration itself. We further measure responsiveness at query time, focusing on the time-to-first-token (TTFT) of the answer, which directly impacts the perceived interactivity of an OEM-VQA assistant.
On QAego4D-Closed, the edge configuration that runson a consumer-grade 8GB GPU achieves an accuracy of $51.76\%\pm0.91$ while satisfying the imposed streaming budgets. 
The best on-premise local configuration reaches $54.40\%\pm0.88$, approaching prior state-of-the-art performance without relying on cloud-based processing.

% Our experimental results on QAEgo4D-Closed demonstrate that lightweight multimodal models can effectively support OEM-VQA in realistic streaming settings. In the edge scenario, our optimal configuration achieves an accuracy of \textcolor{red}{$XX\%$} while satisfying the real-time constraint. These findings highlight both the potential and the limitations of compact models for edge-based episodic memory assistants, providing empirical evidence that purely training-free, lightweight solutions can deliver meaningful performance under realistic constraints.

The contributions of this work are two-fold:
\begin{itemize}
    \item We present the first \emph{systematic study of OEM-VQA under strict real-time constraints on edge hardware}, explicitly targeting privacy-preserving scenarios where cloud offloading is not allowed, and computation must remain local.
    \item We provide an empirical analysis of the \emph{latency--accuracy trade-offs} of lightweight multimodal models on QAEgo4D-Closed, exploring variations in frame rate, resolution, batch size and model size, and identifying operating points and design guidelines for deploying OEM-VQA systems in resource-constrained environments.
\end{itemize}

We hope this research will inform the design of future privacy‑preserving, edge‑based episodic memory and VQA systems.

%% file: Parts/Related_Works.tex
\section{Related Works}

\subsection{Episodic Memory Question Answering}
Episodic Memory (EM) retrieval aims to answer questions about past events observed in egocentric video streams (e.g., ``where did I leave my keys?''). This problem was formalized in Ego4D~\cite{grauman2022ego4dworld3000hours} through the Natural Language Queries (NLQ) task, where models must temporally localize the segment containing the answer.
Beyond temporal grounding, subsequent works introduced open-ended answer generation~\cite{9857465} and multiple-choice formulations such as GroundVQA~\cite{cvpr24_groundvqa}, which reduce the impact of language generation quality by selecting the correct option among distractors.

While these benchmarks were initially studied in offline settings, recent methods target the online regime~\cite{di2025streamingvideoquestionansweringincontext,lando2025oemvqa,shen2024encode,wang2024lifelongmemory}, where the system must process a continuous stream and answer queries without storing the full video. Our work focuses on Online Episodic Memory VQA (OEM-VQA) under strict latency and throughput constraints, explicitly considering privacy-preserving deployments where computation remains on local edge or on-premise hardware.

\subsection{Streaming Multimodal Large Language Models}
Multimodal Large Language Models (MLLMs) have recently achieved strong performance on image and video understanding tasks~\cite{wang2024internvideo2scalingfoundationmodels,li2024llava,zhang2025videollama3frontiermultimodal,geminiteam2024geminifamilyhighlycapable,videollm-online,bai2025qwen3vltechnicalreport}. However, directly extending standard MLLM architectures to streaming video is challenging due to the growth of visual tokens and KV-caches over time. Existing approaches mainly fall into three categories.

\noindent\textbf{KV-cache management.}
Several works enable streaming by controlling the KV-cache footprint. VideoLLM-online~\cite{videollm-online} maintains a rolling cache for long-context streaming, while ReKV~\cite{di2025streamingvideoquestionansweringincontext} retrieves and reuses cached video context at query time by offloading older blocks to disk. Other approaches enforce memory budgets via eviction and stabilization mechanisms, e.g., attention sinks in StreamingVLM~\cite{xu2025streamingvlmrealtimeunderstandinginfinite} and token distillation in InfiniPot~\cite{kim2024infinipotinfinitecontextprocessing}.

\noindent\textbf{Visual token compression.}
A complementary direction reduces redundancy in the visual input. TimeChat-Online~\cite{yao2025timechatonline} drops redundant tokens based on similarity, while Flash-VStream compresses incoming frames into compact state tokens. Video-SALMONN S~\cite{sun2025videosalmonnsstreamingaudiovisual} further explores streaming via continuous adaptation through test-time training.

\noindent\textbf{Textual memory.}
Finally, a lightweight alternative converts visual streams into language-based memories, avoiding long-term storage of visual embeddings. Encode-Store-Retrieve~\cite{shen2024encode} stores language-encoded perception to support semantic retrieval, while LifelongMemory~\cite{wang2024lifelongmemory} organizes long-form egocentric content into textual summaries. EgoLife~\cite{yang2025egolifeegocentriclifeassistant} scales this paradigm with large caption databases and retrieval-augmented generation. Closest to our work, Lando et al.~\cite{lando2025oemvqa} decouple perception and reasoning by generating clip-level textual logs and answering queries solely from text.

In this paper, we adopt the textual memory paradigm and study its feasibility under strict real-time constraints on local hardware, providing a systematic latency--accuracy analysis across edge and enterprise on-premise deployments.

%% file: Parts/Method.tex
\section{\uppercase{Method}}
\label{sec:method}

We address Online Episodic Memory Video Question Answering (OEM-VQA) under a strict \emph{streaming} regime. In this setting, the system must process an egocentric video stream sequentially, adhering to two critical temporal constraints essential for natural user interaction. First, the \textbf{Memory Generation Constraint}: the system must encode each video clip into a memory representation in less time than the clip's duration (real-time throughput), ensuring no backlog is created for subsequent clips. Second, the \textbf{Response Latency Constraint}: upon receiving a user query, the system must retrieve information and generate an answer with minimal latency to maintain fluid conversation.

\begin{figure*}[t] 
    \centering 
    \includegraphics[width=\textwidth]{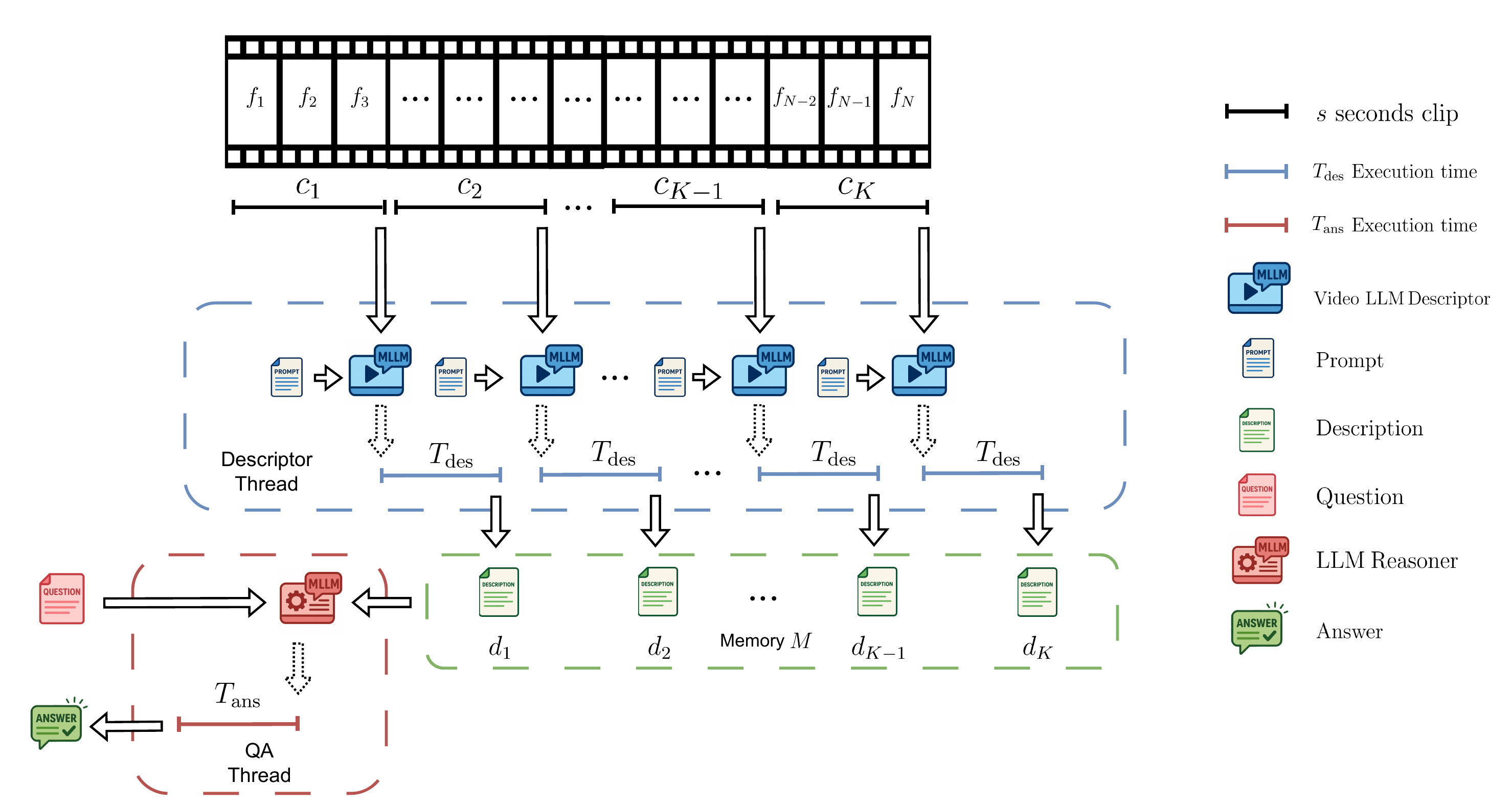} 
    \caption{\textbf{Overview of the Streaming OEM-VQA Framework.} The architecture is organized into two asynchronous threads: \textbf{Descriptor Thread}: Processes handles the continuous streamed video clips ($c_k$) of $s$ seconds. A Video LLM Descriptor generates a textual description ($d_k$) for each clip in execution time $T_{\text{des}}$, incrementally populating the semantic Memory $M$. \textbf{QA Thread}: Activated upon user query, this thread utilizes the stored textual Memory $M$ and a Reasoner model to deduce the Answer in time $T_{\text{ans}}$.} 
    \label{fig:method_overview} 
\end{figure*}

This formulation specifically targets wearable and edge-based assistants such as smart glasses where video is continuously streamed and processed locally without reliance on cloud infrastructure. Our approach adapts the two-stage pipeline proposed in~\cite{lando2025oemvqa}, adhering strictly to these operational bounds. The system incrementally converts incoming video clips into a compact textual memory, within $s$ seconds, through the ``Descriptor Thread'' subprocess and subsequently reasons over this accumulated history through the ``QA Thread'' to answer questions within $t_r$ seconds latency. While conceptually straightforward, this design aligns architectural choices with the rigorous demands of streaming operation and resource-limited deployment. An overview of the pipeline is illustrated in Figure~\ref{fig:method_overview}.

\begin{figure}[t]
    \centering
    \includegraphics[width=\columnwidth]{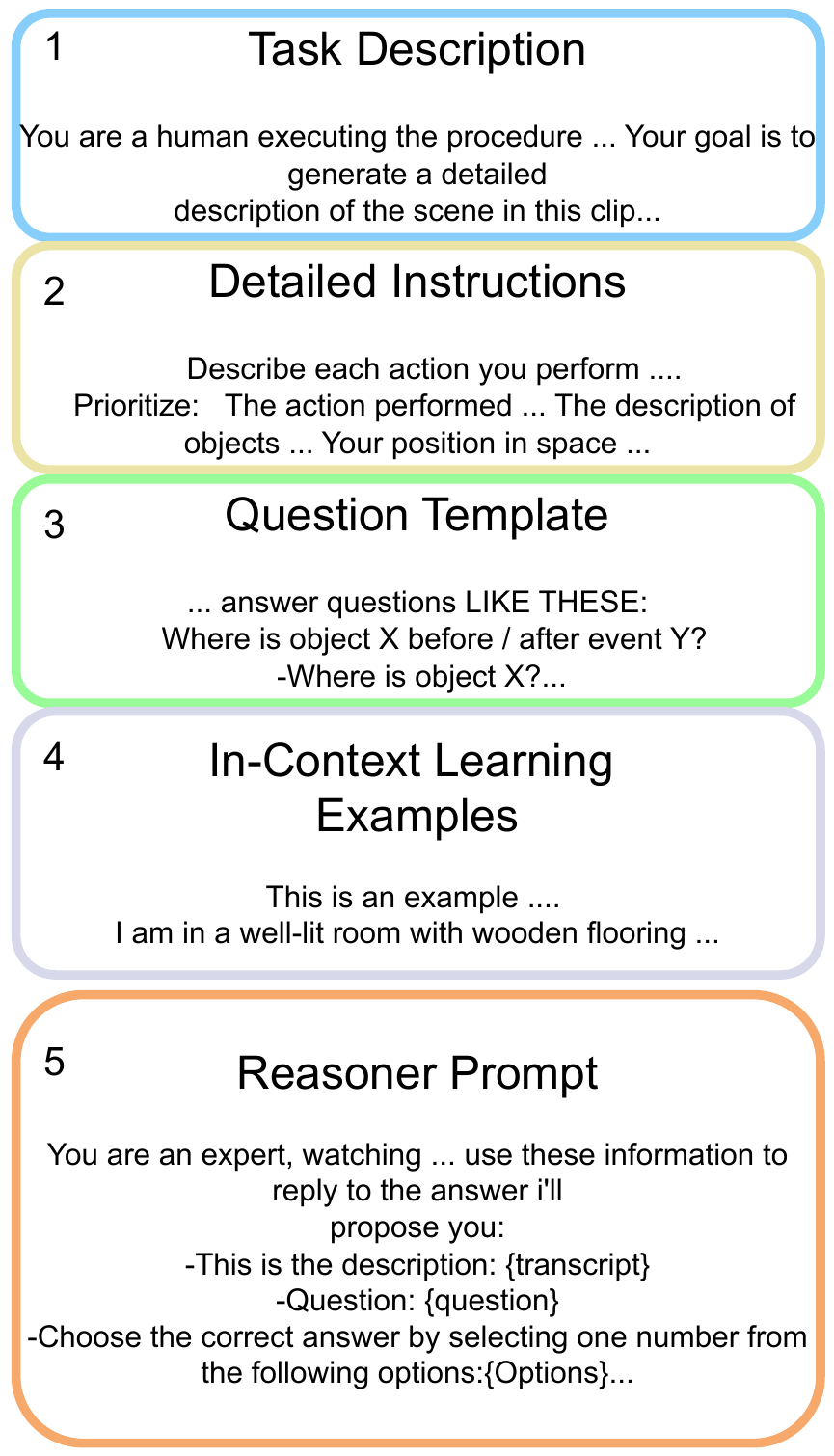} % Replace with your file
    \caption{Overview of the adopted prompting strategy. The Descriptor prompt consists of four main components: \textbf{1) Task Description}: instructs the model on the specific task to perform; \textbf{2) Detailed Instructions}: provides specific guidelines, such as prioritizing actions or spatial positioning; \textbf{3) Question Template}: primes the model with potential future questions; and \textbf{4) In-Context Learning Examples}: provides a full clip description example to encourage adherence to output guidelines. Additionally, the \textbf{5) Reasoner Prompt} is used at query time, providing the model with the question, candidate answers, and the accumulated memory history.}
    \label{fig:Prompt}
\end{figure}

\subsection{Descriptor Thread}
\label{subsec:memory_construction}

Let an egocentric video stream be represented as a sequence of frames
\begin{equation}
v = (f_1, f_2, \dots, f_N),    
\end{equation}
where each frame $f_i \in \mathbb{R}^{H \times W \times 3}$. The stream is processed into a sequence of non-overlapping clips
\begin{equation}
C = (c_1, c_2, \dots, c_K),
\end{equation}
where each clip $c_k$ spans a fixed temporal duration of $s$ seconds. Clips are processed sequentially as soon as they are observed, without access to future frames.

Each clip $c_k$ is processed by the \emph{Descriptor Thread}, an independent subprocess that manages a lightweight Multimodal Large Language Model (MLLM) and generates a textual description $d_k$ summarizing the visual content from a first-person perspective. 
To satisfy real-time streaming during memory construction, the descriptor must complete within the clip duration:
\begin{equation}
T_{\text{des}}(c_k) < s.
\label{eq:streaming_desc}
\end{equation}

The textual memory $M$ is defined as the ordered sequence of clip-level descriptions:
\begin{equation}
M = (d_1, d_2, \dots, d_K).
\label{eq:text_memory}
\end{equation}
This memory grows incrementally over time and serves as a persistent, human-readable representation of the observed video. Importantly, raw video frames are discarded after description generation, making the memory lightweight and privacy-preserving.

\subsection{QA Thread}
\label{subsec:reasoning}

Whenever the user formulates a question $q$, the \emph{QA Thread} responds by relying on the textual memory $M$, without re-accessing the original video stream. 
Given a question $q$, the task is to infer the correct answer by reasoning solely over $M$ under an application-level responsiveness budget $t_r$:
\begin{equation}
T_{\text{ans}}(M,q) < t_r.
\label{eq:streaming_qa}
\end{equation}

We consider the closed-ended OEM-VQA setting adopted in QAEgo4D-Closed~\cite{cvpr24_groundvqa}, where each query $q$ is associated with four candidate answers and the output is a discrete choice:
\begin{equation}
a \in \{A, B, C, D\}.
\end{equation}

The \emph{reasoning module} (within the QA Thread) receives the concatenation of the memory $M$, the question $q$, and the candidate answers as input. No visual retrieval, filtering, or re-encoding is performed at query time; the entire reasoning process operates in the textual domain. Formally, the predicted answer is produced as follows:
\begin{equation}
a = R(M, q),
\end{equation}
where $R(\cdot)$ denotes the reasoning model.

\subsection{Prompt Design}
\label{subsec:prompt_design}

Effective prompting is critical for both the description and reasoning phases of the pipeline. We adopt a structured design (Figure~\ref{fig:Prompt}) to ensure the outputs of the MLLM align with the requirements of the streaming task. 
To guide the descriptor toward generating information useful for downstream episodic queries, the prompt instructs the model to describe the scene in a first-person narrative and explicitly encourages the inclusion of details relevant to typical episodic memory questions. Specifically, after providing general and detailed instructions, we incorporate \emph{template questions} derived from the NLQ annotation guidelines of Ego4D~\cite{grauman2022ego4dworld3000hours}, such as object locations and recent actions, which act as a form of soft supervision.
Based on findings from prior ablation studies~\cite{lando2025oemvqa}, we do not include the previous clip description as contextual input when generating $d_k$. This choice avoids error accumulation across clips and keeps the prompt length bounded. All clips are therefore described independently using the same prompt structure. Finally, for the reasoning model, we employ a distinct Reasoner Prompt. This prompt concatenates the entire accumulated textual memory $M$—serving as the context—with the specific user question $q$ and the set of candidate answers. The model is strictly instructed to analyze the provided textual history and select the correct option from the choices, without generating extraneous reasoning steps.

%\subsection{Streaming and Deployment Considerations}
%\label{subsec:streaming_constraints}

%The proposed pipeline is explicitly designed to operate under streaming constraints typical of edge-based egocentric assistants. First, memory construction is performed online, with each clip processed independently and discarded after description generation. Second, the textual memory is compact, growing linearly with time at a rate of only a few kilobytes per minute. Third, query-time reasoning is performed with low latency ($T_\text{ans}$), which we operationally measure as the time-to-first-token (TTFT) to ensure immediate responsiveness.

%Crucially, the entire pipeline is intended to run on local hardware—ranging from consumer-level devices to more capable on-premise servers—without relying on cloud computation. This makes the approach suitable for privacy-sensitive scenarios where continuous video streams cannot be transmitted off-device, such as wearable smart glasses or assistive systems deployed in controlled environments.

%% file: Parts/Results.tex
\section{\uppercase{Experimental Settings}}
\label{sec:experiments}

All experiments were conducted on the \textsc{QAEgo4D-Closed} benchmark~\cite{cvpr24_groundvqa}, which consists of 500 multiple-choice questions (four candidate answers) defined over egocentric videos from Ego4D~\cite{grauman2022ego4dworld3000hours}. Since the task is closed-ended question answering, we report \emph{accuracy} (\%) as the primary evaluation metric.

\subsection{\uppercase{Deployment Scenarios}}
\label{subsec:deployment}

We consider two privacy-preserving deployment regimes in which raw video never leaves the local infrastructure. The first represents a consumer-grade edge setting (e.g., smart glasses streaming to a nearby personal device) equipped with a single GPU. The second represents an enterprise setting with a more capable local GPU server, suitable for institutions operating under strict privacy constraints (e.g., hospitals or care facilities), while still avoiding any cloud-based processing. In our experiments, these scenarios are instantiated with an NVIDIA RTX~3070 (8GB) and an NVIDIA L40S (48GB), respectively.

\subsection{\uppercase{Streaming Constraint}}
\label{subsec:streaming_constraint}

We adopt the streaming budgets formalized in Eqs.~\eqref{eq:streaming_desc}--\eqref{eq:streaming_qa}. 
In this work we set $s=15$s, matching the best-performing streaming configuration adopted as starting point in~\cite{lando2025oemvqa}. 
We set $t_r=1$s as an upper bound for our proposed pipeline, since within a second to reply a user question the system it's perceived as streaming and not as a delayed one. %We evaluate query-time responsiveness via Time To First Token (TTFT) measurements (Section~\ref{subsec:reasoner_ttft}).

\subsection{\uppercase{Models and Setup}}
\label{subsec:models_setup}

All configurations in this study employ models from the Qwen3-VL family~\cite{qwen2025qwen25technicalreport}. We exclusively use \emph{Instruct} variants (rather than \emph{Thinking}) because explicit reasoning traces introduce additional output tokens and latency, making it unsuitable for satisfying the real-time constraint in Eq.~\eqref{eq:streaming_qa}. 
All experiments are run with FlashAttention-2~\cite{dao2023flashattention} reducing attention overhead and improve throughput.
For both captioning and question answering, runtimes are measured as the wall-clock difference between timestamps immediately before and after the model's \texttt{generate()} call, averaged over 10 runs.

% We follow the training-free OEM-VQA paradigm introduced in~\cite{lando2025oemvqa}: a multimodal model produces clip-level textual descriptions which are appended to a growing memory, and the question-answering module selects one option among four candidates by reading the memory.

% Regarding prompting, we adopt the same descriptor prompt design as~\cite{lando2025oemvqa}, including template questions derived from Ego4D NLQ guidelines~\cite{grauman2022ego4dworld3000hours}. We do not provide previous-clip descriptions as context when generating the current clip description, and we keep the clip length fixed to 15 seconds.

\begin{table}[t]
\caption{Configuration selection on the enterprise setting (L40S, 48GB).
The model is Qwen3-VL. We report mean$\pm$std for time per clip, generation throughput,
and peak GPU memory. The \textbf{bold} row denotes the selected configuration.}
\label{tab:grid_l40s}
\centering
\scriptsize
\setlength{\tabcolsep}{3pt}
\renewcommand{\arraystretch}{0.92}

\begin{tabular}{lccccc}
\hline
Size & BS & 
\begin{tabular}{@{}c@{}}Time/Clip\\(s)$\downarrow$\end{tabular} &
\begin{tabular}{@{}c@{}}Tok/s\\$\uparrow$\end{tabular} &
\begin{tabular}{@{}c@{}}Peak Mem\\(GB)$\downarrow$\end{tabular} \\
\hline
2B & 1 & 7.30$\pm$1.62  & 32.68$\pm$3.91 & 6.22$\pm$0.07 \\
2B & 2 & 5.89$\pm$1.04  & 50.25$\pm$11.28 & 8.51$\pm$0.12 \\
2B & 4 & 7.10$\pm$0.86  & 48.24$\pm$0.81 & 13.24$\pm$0.01 \\
\textbf{4B} & \textbf{1} & \textbf{9.76$\pm$2.12} & \textbf{29.71$\pm$4.98} & \textbf{10.73$\pm$0.03} \\
4B & 2 & 7.18$\pm$0.71  & 35.25$\pm$2.07 & 13.81$\pm$0.01 \\
4B & 4 & 6.19$\pm$0.75  & 54.40$\pm$5.59 & 19.96$\pm$0.02 \\
8B & 1 & 13.42$\pm$3.35 & 20.47$\pm$1.77 & 19.74$\pm$0.02 \\
8B & 2 & 9.31$\pm$1.39  & 31.42$\pm$0.86 & 23.03$\pm$0.01 \\
8B & 4 & 13.17$\pm$2.24 & 30.90$\pm$4.68 & 30.13$\pm$0.02 \\
\hline
\end{tabular}
\end{table}

\subsection{\uppercase{Configuration Selection Under Real-Time Constraints}}
\label{subsec:config_selection}

The key experimental question of this paper is how to select configurations that satisfy the streaming budgets in Eqs.~\eqref{eq:streaming_desc}--\ref{eq:streaming_qa} while remaining accurate and feasible under edge deployment constraints. We therefore perform systematic configuration selection studies in each deployment regime, and subsequently use the selected configurations for end-to-end accuracy evaluation.

%\subsubsection{\uppercase{consumer-grade Edge Configuration (8GB GPU)}}
%\label{subsubsec:grid_3070}

On the consumer-grade edge setting, we perform a grid search over video ingestion and batching parameters to identify configurations that can generate one 15s clip description in less than 15 seconds. Specifically, we vary: (i) input frame rate (fps), (ii) input resolution, (iii) batch size, and (iv) quantization mode. For each configuration, we report the mean and standard deviation of (a) time per clip, (b) tokens per second, and (c) peak GPU memory usage. The full ablation is summarized in Table~\ref{tab:grid_3070}.

Among the configurations satisfying Eq.~\eqref{eq:streaming_desc}, we select the operating point that maximizes input fidelity (higher fps and resolution) while remaining within the real-time budget and leaving sufficient GPU memory headroom for query answering. As shown in Table~\ref{tab:grid_3070}, this corresponds to fps=2, native resolution, and batch size=1 under the non-quantized setting.

%\subsubsection{\uppercase{enterprise Configuration (48GB GPU)}}
%\label{subsubsec:grid_l40s}

On the enterprise setting, the larger memory budget enables scaling in model size and batch size. We therefore perform a second configuration selection study where we vary the Qwen3-VL model size and the batch size, while keeping the streaming requirement in Eq.~\eqref{eq:streaming_desc}. Table~\ref{tab:grid_l40s} reports the same set of runtime and resource statistics as in the consumer-grade setting. This study allows identifying the best enterprise operating point under real-time constraints. On the enterprise GPU, Table~\ref{tab:grid_l40s} shows that the larger memory budget enables scaling the model size while still meeting the streaming constraint. We therefore select the largest model that achieves the lowest time/clip within the feasible batch-size range (Qwen3-VL-8B with batch size 2 in our study).

\begin{table}[t]
\caption{Configuration selection on the enterprise setting (L40S, 48GB).
The model is Qwen3-VL. We report mean$\pm$std for time per clip, throughput,
and peak GPU memory. The \textbf{bold} row denotes the selected configuration.}
\label{tab:grid_l40s}
\centering
\small
\setlength{\tabcolsep}{3pt}
\renewcommand{\arraystretch}{1.05}

\begin{tabularx}{\linewidth}{@{}l c >{\centering\arraybackslash}X
                                >{\centering\arraybackslash}X
                                >{\centering\arraybackslash}X@{}}
\hline
Size & BS & Time/Clip (s)$\downarrow$ & Tok/s$\uparrow$ & Peak Mem. (GB)$\downarrow$ \\
\hline
2B & 1 & 7.30$\pm$1.62  & 32.68$\pm$3.91  & 6.22$\pm$0.07 \\
2B & 2 & 5.89$\pm$1.04  & 50.25$\pm$11.28 & 8.51$\pm$0.12 \\
2B & 4 & 7.10$\pm$0.86  & 48.24$\pm$0.81  & 13.24$\pm$0.01 \\
\textbf{4B} & \textbf{1} & \textbf{9.76$\pm$2.12} & \textbf{29.71$\pm$4.98} & \textbf{10.73$\pm$0.03} \\
4B & 2 & 7.18$\pm$0.71  & 35.25$\pm$2.07  & 13.81$\pm$0.01 \\
4B & 4 & 6.19$\pm$0.75  & 54.40$\pm$5.59  & 19.96$\pm$0.02 \\
8B & 1 & 13.42$\pm$3.35 & 20.47$\pm$1.77  & 19.74$\pm$0.02 \\
8B & 2 & 9.31$\pm$1.39  & 31.42$\pm$0.86  & 23.03$\pm$0.01 \\
8B & 4 & 13.17$\pm$2.24 & 30.90$\pm$4.68  & 30.13$\pm$0.02 \\
\hline
\end{tabularx}
\end{table}

\begin{table}[t]
\caption{Query-time responsiveness measured as time-to-first-token (TTFT, mean$\pm$std) under different deployment settings. Larger models exceed the available memory on the consumer-grade GPU and result in out-of-memory (OOM) errors.}
\label{tab:ttft_combined}
\centering
\small
\setlength{\tabcolsep}{3pt}
\renewcommand{\arraystretch}{1.05}
\begin{tabular}{@{}p{0.27\linewidth}p{0.37\linewidth}c@{}}
\hline
GPU & Model & TTFT (s)$\downarrow$ \\
\hline
\multirow{3}{=}{RTX 3070 (8GB)}
 & Qwen3-VL-2B & 0.41$\pm$0.27 \\
 & Qwen3-VL-4B & OOM \\
 & Qwen3-VL-8B & OOM \\
\hline
\multirow{3}{=}{L40S (48GB)}
 & Qwen3-VL-2B & 0.20$\pm$0.14 \\
 & Qwen3-VL-4B & 0.49$\pm$0.31 \\
 & Qwen3-VL-8B & 0.88$\pm$0.57 \\
\hline
\end{tabular}
\end{table}

\subsection{\uppercase{Query-Time Responsiveness via Time-To-First-Token}}
\label{subsec:reasoner_ttft}

Beyond streaming memory construction, an OEM-VQA assistant must respond promptly to user queries. This directly relates to the query-time budget in Eq.~\ref{eq:streaming_qa}.
To quantify query-time responsiveness, we measure \emph{time-to-first-token} (TTFT) as the wall-clock time between the start of generation and the production of the first output token. We implement this measurement by setting \texttt{max\_new\_tokens=1}, enforcing the model to output the answer option as the first token.
We evaluate the model only on the first produced token since the perceived delay of a streaming application is not affected by ``How much" the model says but by ``how long" the model takes to start its reply.

 Results are summarized in Table~\ref{tab:ttft_combined}. On the consumer-grade 8GB GPU, only the 2B model can be evaluated, while larger variants (4B and 8B) exceed memory limits (OOM). On the enterprise GPU, all model sizes are feasible, but TTFT increases with model capacity, reflecting a clear latency--accuracy trade-off: larger models improve end-to-end accuracy (Section~\ref{subsec:main_results}) at the cost of higher query-time delay. Notably, the selected configurations are compatible with the responsiveness budget in Eq.~\eqref{eq:streaming_qa} on average.

\begin{table}[t]
\caption{Main results on QAego4D-Closed test set (500 questions). 
All configurations use Qwen3-VL Instruct models (2B: Qwen3-VL-2B-Instruct, 8B: Qwen3-VL-8B-Instruct). 
Accuracy is reported as mean$\pm$std over 10 runs with different seeds. ``On Edge'' indicates whether the full pipeline can run end-to-end on an RTX 3070 (8GB).}
\label{tab:main_results_4configs}
\centering
\small
\setlength{\tabcolsep}{3pt}
\renewcommand{\arraystretch}{1.05}
\begin{tabular}{@{}c c c c@{}}
\hline
Descriptor & Reasoner & On Edge & Accuracy (\%)$\uparrow$ \\
\hline
2B & 2B & Yes & 51.76 $\pm$ 0.91 \\
2B & 8B & No  & 52.54 $\pm$ 1.11 \\
8B & 2B & No  & 51.32 $\pm$ 1.33 \\
8B & 8B & No  & 54.40 $\pm$ 0.88 \\
\hline
\end{tabular}
\end{table}

\begin{table*}[t]
\caption{Comparison with state-of-the-art methods on QAego4D-Closed in terms of accuracy. 
We report mean$\pm$std only for our runs (10 seeds).}
\label{tab:sota_comparison_edge_onprem}
\centering
\setlength{\tabcolsep}{10pt}
\begin{tabular}{l c}
\hline
Method & Accuracy (\%)$\uparrow$ \\
\hline
Ground VQA~\cite{cvpr24_groundvqa} & 48.70 \\
RekV-LLavaOneVision 0.5~\cite{di2025streamingvideoquestionansweringincontext} & 50.00 \\
RekV-LLaVaOneVision 7B~\cite{di2025streamingvideoquestionansweringincontext} & 56.00 \\
LLaVaOneVision-based~\cite{lando2025oemvqa} & 51.88 \\
Gemini-based~\cite{lando2025oemvqa} & 56.00 \\
\hline
Ours (Edge-best: 2B Desc + 2B Reasoner) & 51.76 $\pm$ 0.91 \\
Ours (On-prem-best: 8B Desc + 8B Reasoner) & 54.40 $\pm$ 0.88 \\
\hline
\end{tabular}
\end{table*}

\subsection{\uppercase{Main Results}}
\label{subsec:main_results}

Table~\ref{tab:main_results_4configs} reports the accuracy of the four descriptor--reasoner combinations obtained by mixing edge-generated descriptions (2B) and enterprise-generated descriptions (8B) with different query-time model sizes. While hybrid combinations are included for completeness, our target scenario assumes limited resources and minimal system complexity; hence, we focus on configurations where a \emph{single} model is used for both description generation and query-time answering, avoiding the overhead of loading and swapping multiple models at query time.

Under the edge deployment regime, the 2B+2B configuration is the only end-to-end setup that fits the consumer-grade 8GB GPU while satisfying the descriptor streaming budget (Eq.~\ref{eq:streaming_desc}) and maintaining interactive query-time responsiveness (Eq.~\ref{eq:streaming_qa}, measured via TTFT in Section~\ref{subsec:reasoner_ttft}).

Scaling to the enterprise regime, the 8B+8B configuration yields the best overall performance (54.40\%), highlighting the expected trade-off between latency and accuracy: larger models tend to improve answer quality, but incur higher TTFT and require stronger local hardware.

Table~\ref{tab:sota_comparison_edge_onprem} compares two deployment configurations (edge-best and enterprise-best) against prior work. Our results remain near state of the art while operating under streaming constraints and privacy-preserving local processing, showing that real-time OEM-VQA is feasible without cloud dependence.

%% file: Parts/Conclusion.tex
\section{CONCLUSION}
We tackled the challenge of deploying privacy-preserving, real-time Online Episodic Memory Video Question Answering (OEM-VQA) directly on edge hardware. %We addressed the challenge of limited resources by imposing strict streaming constraints, ensuring that memory generation operates faster than real-time while satisfying low-latency query requirements. To achieve this, we proposed a decoupled architecture comprising two asynchronous threads: a Descriptor Thread that continuously converts video into a lightweight textual memory, and a QA Thread that reasons over this episodic history upon user demand. 
Our experiments confirm that lightweight Multimodal Large Language Models can effectively address this task within strict resource constraints. Specifically, an end-to-end configuration on a consumer-grade 8GB GPU achieved an accuracy of 51.76\% with a Time-To-First-Token (TTFT) of 0.41s. Scaling to a local enterprise-grade server increased accuracy to 54.40\% (0.88s TTFT), approaching the performance of cloud-based solutions without compromising privacy. We believe this study offers critical insights for the design of future edge-based systems and paves the way for further research on autonomous wearable assistants.

% Ack
\paragraph{Acknowledgements}
This research has been funded by the European Union - Next Generation EU, Mission 4 Component 1 CUP E53D23016240001 - Project PRIN 2022 PNRR TEAM and by the project  PIACERI - PIAno di inCEntivi per la Ricerca di Ateneo 2024/2026 — Linea di Intervento i ``Progetti di ricerca collaborativa''. We acknowledge ISCRA for awarding this project access to the LEONARDO supercomputer, owned by the EuroHPC Joint Undertaking, hosted by CINECA (Italy).